\documentclass[runningheads]{llncs}
\usepackage[T1]{fontenc}
\usepackage{booktabs}  % 提供 \toprule, \midrule, \bottomrule
\usepackage{graphicx}  % 提供 \resizebox
\usepackage{graphicx,verbatim}
\usepackage{amsmath}
\usepackage{amssymb}
\usepackage{amsfonts}
\usepackage{marvosym}
\usepackage{multirow}
\usepackage{booktabs}
\usepackage{multirow}
\usepackage{cite}
\usepackage{enumitem}
\usepackage{url}
\usepackage{placeins}
\usepackage[hidelinks]{hyperref}
\newcommand{\equalcontrib}{\textsuperscript{\dag}}
\newcommand{\corrauth}{\textsuperscript{(\Letter)}}
\begin{document}
\title{Real-Time Hardware-Free HIFU Interference Suppression via Teacher-Student Diffusion Framework}
\begin{comment}  %% Removed for anonymized MICCAI submission
\author{First Author\inst{1}\orcidID{0000-1111-2222-3333} \and
Second Author\inst{2,3}\orcidID{1111-2222-3333-4444} \and
Third Author\inst{3}\orcidID{2222--3333-4444-5555}}
%
\authorrunning{F. Author et al.}
\institute{Princeton University, Princeton NJ 08544, USA \and
Springer Heidelberg, Tiergartenstr. 17, 69121 Heidelberg, Germany
\email{lncs@springer.com}\\
\url{http://www.springer.com/gp/computer-science/lncs} \and
ABC Institute, Rupert-Karls-University Heidelberg, Heidelberg, Germany\\
\email{\{abc,lncs\}@uni-heidelberg.de}}

\end{comment}

\author{Dejia Cai\equalcontrib\inst{1} \and
Ali Abdollahi\equalcontrib\inst{1} \and
Xi Wang\inst{1} \and
Kun Yang\inst{2} \and
Zhaohui Guo\inst{3} \and
Xiaowei Zhou\inst{2}\corrauth \and
Hao Chen\inst{1,4,5,6,7}\corrauth}
% index{Cai, Dejia}
% index{Abdollahi, Ali}
% index{Wang, Xi}
% index{Yang, Kun}
% index{Guo, Zhaohui}
% index{Zhou, Xiaowei}
% index{Chen, Hao}
\authorrunning{D. Cai et al.}
\institute{Department of Computer Science and Engineering, The Hong Kong University of Science and Technology, Hong Kong SAR, China\\
\email{jhc@cse.ust.hk}
\and
State Key Laboratory of Ultrasound Engineering in Medicine, Chongqing Medical University, Chongqing 400016, China\\
\email{zhou.xiaowei@cqmu.edu.cn}
\and
School of Microelectronics, Tianjin University, Tianjin 300072, China
\and
Department of Chemical and Biological Engineering, The Hong Kong University of Science and Technology, Hong Kong SAR, China
\and
Division of Life Science, The Hong Kong University of Science and Technology, Hong Kong SAR, China
\and
HKUST Shenzhen-Hong Kong Collaborative Innovation Research Institute, The Hong Kong University of Science and Technology, Futian, Shenzhen, China
\and
State Key Laboratory of Nervous System Disorders, The Hong Kong University of Science and Technology, Hong Kong SAR, China}
  
\maketitle              % typeset the header of the 
\begingroup
\renewcommand{\thefootnote}{\dag}
\footnotetext{Equal contribution.}
\endgroup

\begin{abstract}

High-Intensity Focused Ultrasound (HIFU) is a non-invasive therapy, yet its safety is often degraded by severe acoustic interference during continuous ultrasound guidance. Existing suppression methods heavily rely on proprietary raw Radio-Frequency (RF) data or hardware synchronization, limiting their clinical utility and preventing real-time implementation. We propose Manifold-Constrained Hyper-Connections Diffusion (mHC-Diff), an image-domain diffusion framework for real-time interference suppression without specialized hardware synchronization, disentangling complex interference from anatomical structures while ensuring high reconstruction fidelity. To achieve clinical real-time application, mHC-Diff first learns an anatomy-aware prior with a high-fidelity multi-step diffusion Teacher, then distills it into a one-step Student. On a clinically representative dataset spanning diverse therapeutic scenarios, mHC-Diff achieves 26.65 dB PSNR and real-time inference ($\sim$20 FPS) on a single NVIDIA RTX 4090, a $\sim$6.8$\times$ speedup over iterative diffusion baselines such as HIFU-Diff. These results suggest a practical route to deployable, hardware-free interference suppression for ultrasound-guided HIFU interventions. Code is available at \url{https://github.com/caidejia/HIFU-mHC-Diff}.

\keywords{HIFU \and Interference Suppression \and Diffusion Models \and Manifold Learning \and Model Distillation \and Real-time Imaging.}
% Authors must provide keywords and are not allowed to remove this Keyword section.

\end{abstract}

\section{Introduction}

High-Intensity Focused Ultrasound (HIFU) has emerged as a widely adopted non-invasive therapeutic modality~\cite{bond2017safety,chen2022ultrasound,hynynen1993mri}, used to treat various pathologies, including uterine fibroids~\cite{dou2024long}, breast fibroadenomas~\cite{imankulov2018hifu}, and adenomyosis~\cite{liu2025machine}. Although HIFU surpasses conventional surgical interventions in patient recovery and tissue preservation, its clinical success depends fundamentally on high-precision, real-time monitoring. Ultrasound imaging provides cost-effective, dynamic intra-operative guidance. However, the intense acoustic waves emitted by the therapeutic HIFU transducer interfere with the imaging pulses, degrading the signal-to-noise ratio of the acquired images. To mitigate this, current protocols rely on a ``stop-and-shoot'' approach, pausing sonication during imaging. This compromise can prolong procedure times, reduce treatment efficacy, and increase clinical risks associated with unguided sonication~\cite{imankulov2018hifu}.

To enable concurrent ultrasound monitoring during continuous sonication, various signal processing and deep learning algorithms have been propose~\cite{yang2024frequency,shen2022golay,song2014correspondence,shen2024ultrasound,payen2023passive,lee2021fus,yang2024suppressing,cai2024novel}. Traditional methods, like FRPCA ~\cite{yang2024frequency} or encoded excitations ~\cite{shen2022golay,song2014correspondence,shen2024ultrasound}, often lack clinical generalization or impose high computational overhead. Recently, deep learning architectures, ranging from FUS-Net~\cite{lee2021fus} to diffusion-based HIFU-Diff~\cite{cai2024suppressing}, have emerged as powerful alternatives for robust interference suppression. However, these methods share a major bottleneck: they predominantly rely on accessing raw Radio-Frequency (RF) data or demand complex hardware synchronization. In clinical practice, raw RF data is typically proprietary and restricted on commercial scanners. Furthermore, processing raw RF signals is computationally demanding, complicating real-time deployment and limiting clinical versatility.

\begin{figure}[t] 
    \centering  
    \includegraphics[width=0.8\linewidth]{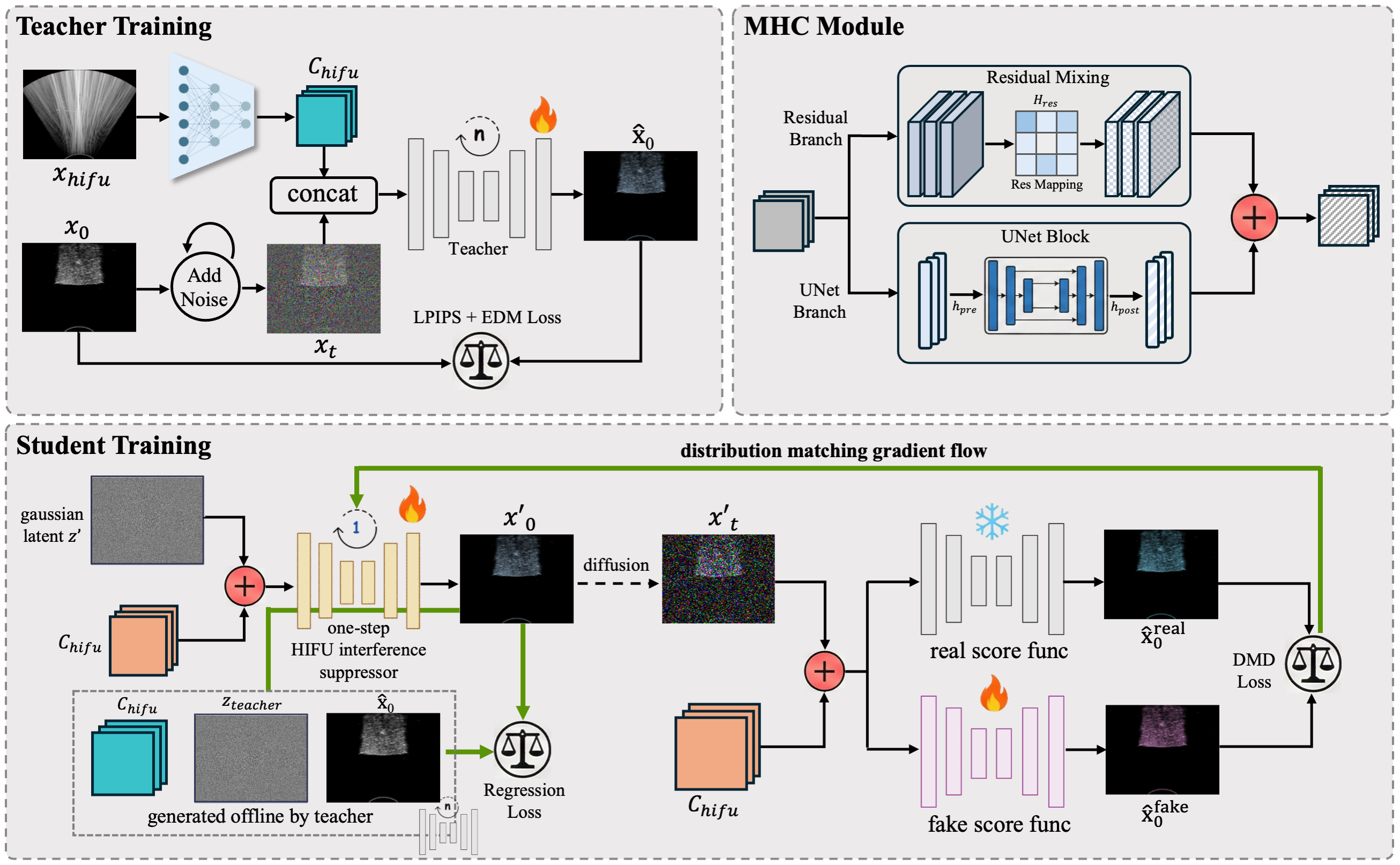} 
    \caption{\textbf{The mHC-Diff framework.} A two-stage pipeline for real-time interference suppression: (i) prior learning via a diffusion-based Teacher, and (ii) knowledge distillation into a single-step Student for real-time inference.}
    \label{fig:framework}
\end{figure}

Consequently, there is an urgent need for a hardware-free, image-domain solution operating on standard B-mode images. We conceptualize the restoration of corrupted images as mapping contaminated observations back into the clean anatomical domain. While diffusion models~\cite{ho2020denoising,dhariwal2021diffusion,esser2021taming,zhang2021designing} effectively capture such complex manifolds~\cite{saharia2022image,niu2024acdmsr}, translating them to real-time HIFU guidance is challenged by slow iterative sampling and the limited capacity of existing U-Nets under extreme interference. To bridge this gap, we propose mHC-Diff, an image-domain diffusion framework that enables high-fidelity prior learning and facilitates knowledge distillation for single-step, real-time acceleration. Specifically, our framework leverages the mHC-UNet architecture, which incorporates manifold-constrained Hyper-Connections (mHC) to expand model capacity via multi-stream residual pathways with minimal computational overhead. By enforcing manifold constraints inspired by~\cite{xie2025mhc}, this design stably disentangles intense acoustic artifacts from anatomical structures, ensuring high structural fidelity without destructive signal loss. Our primary contributions are:

\begin{itemize}
\item We introduce mHC-Diff, an image-domain diffusion framework for HIFU interference suppression operating on standard B-mode inputs, obviating proprietary raw RF access and specialized hardware synchronization.
\item We propose the mHC-UNet architecture and a two-stage knowledge distillation strategy that disentangles interference from anatomy. This approach achieves superior restoration (26.65 dB PSNR), while accelerating inference to real-time clinical standards ($\sim$20 FPS).
\item We constructed a large-scale clinical equivalent HIFU interference suppression dataset comprising 17,464 image pairs from ex vivo and in vivo tissues, covering three typical imaging modes and four HIFU irradiation intensities to support the development of HIFU interference suppression methods.
\end{itemize}  
\section{Methods}

mHC-Diff is a framework for real-time restoration of ultrasound images corrupted by high-intensity focused ultrasound (HIFU) interference, as show in Fig.~\ref{fig:framework}. It operates in two stages: (i) manifold prior acquisition, where a conditional diffusion Teacher is trained to capture anatomy-consistent clean anatomical distributions; and (ii) where the learned prior is distilled into a single-step Student for real-time inference. Both stages utilize mHC-UNet (Sec.~\ref{sec:mhc_unet}) to disentangle acoustic interference from anatomical structures.

\subsection{Network Architecture: Concat-Conditioned mHC-UNet}
\label{sec:mhc_unet}

We propose a shared denoiser backbone for the Teacher, Student, and critic: an mHC-UNet augmented by manifold constraints inspired by~\cite{xie2025mhc}.

\textbf{Conditioning and network input.}
Given a corrupted frame $x_{\mathrm{hifu}}$ and a noisy target $x$, We extract spatial conditioning features $C_{\mathrm{hifu}} = E_{\psi}(x_{\mathrm{hifu}})$, where $E_{\psi}$ denotes the HIFU-aware conditioning encoder. We concatenate them as $u = [x, C_{\mathrm{hifu}}]$ and feed $u$ into the U-Net stem, where the initial feature map is explicitly broadcast into $n=4$ parallel latent streams. The noise level $\sigma$ is injected into all blocks as a global conditional signal, and the $n$ streams are aggregated via feature averaging prior to the final prediction layer.

\textbf{Multi-stream residual pathways with mHC routing.}
Following the paradigm in~\cite{xie2025mhc}, our architecture maintains $n=4$ parallel streams exclusively along the residual pathways throughout the UNet structure, applying shared-weight projections per stream to improve parameter efficiency. 

Cross-stream interaction and integration with standard U-Net blocks are achieved via the adopted mHC module. Because the core U-Net operator $\mathcal{F}(\cdot, \sigma)$ requires a single unified input, the module predicts non-negative read and write gates, $h_{\mathrm{pre}}(p), h_{\mathrm{post}}(p) \in \mathbb{R}^{n}_{+}$, alongside a residual stream-mixing matrix $H_{\mathrm{res}}(p) \in \mathbb{R}^{n\times n}$. At spatial location $p$, the read gates aggregate the $n$ residual streams $x_k(p)$ into a unified representation $r(p)$ for the U-Net block. The write gates then distribute the resulting update $u(p)$ back to the individual streams, which are concurrently mixed by $H_{\mathrm{res}}$:
\begin{equation}
\begin{aligned}
r(p) &= \textstyle\sum_{k=1}^{n} h_{\mathrm{pre},k}(p)\,x_k(p), \qquad u(p) = \mathcal{F}\!\left(r(p),\,\sigma\right), \\
x^{\mathrm{out}}_i(p) &= \textstyle\sum_{j=1}^{n} H_{\mathrm{res},ij}(p)\,x_j(p) \;+\; h_{\mathrm{post},i}(p)\,u(p), \qquad \forall i \in \{1,\dots,n\}.
\end{aligned}
\end{equation}
To ensure stability and identity mapping, we enforce manifold constraints: $H_{\mathrm{res}}$ is Sinkhorn-normalized to the Birkhoff polytope to conserve feature magnitude, while sigmoid-activated gates prevent destructive signal cancellation.

\subsection{Teacher Prior Learning and Single-Step Distillation}\label{sec:dmd}
To balance restoration fidelity and real-time efficiency, we
propose a two-stage strategy: first, capturing an anatomy-aware prior via an Elucidated
Diffusion Model (EDM)~\cite{karras2022elucidating}, and then accelerating inference through Distribution
Matching Distillation (DMD)~\cite{yin2024one}.

\textbf{Stage I: Teacher Optimization via Elucidated Diffusion Model.} We first train a Teacher $\mu_{\mathrm{Teacher}}$ to capture the clean anatomical manifold. Following EDM, we corrupt the clean target $x_0$ with noise level $\hat{\sigma}$ to obtain $x_t = x_0 + \hat{\sigma} \epsilon$. To preserve fine acoustic textures, the Teacher is optimized to reconstruct $x_0$ conditioned on frozen $C_{\mathrm{hifu}}$ via a perceptual-augmented loss:
\begin{equation}
\mathcal{L}_{\mathrm{Teacher}} = \mathbb{E}_{\hat{\sigma}} \!\left[ \hat{\sigma}^{-2} \left\| \hat{x}_0 - x_0 \right\|_2^2 + \lambda_{\mathrm{perc}}\,\mathrm{LPIPS}(\hat{x}_0, x_0) \right],
\end{equation}
where $\hat{x}_0 = \mu_{\mathrm{Teacher}}([x_t, C_{\mathrm{hifu}}], \hat{\sigma})$ is the reconstructed image, $\hat{\sigma}$ is the standard deviation of injected noise. Once fully trained, the Teacher is frozen as a reference denoiser $\mu_{\mathrm{real}}$ for the subsequent phase.

\textbf{Stage II: One-Step Distillation and Denoising Field Matching.} We distill $\mu_{\mathrm{real}}$ into a one-step Student generator $G_{\phi}$ via DMD, involving the Student $G_{\phi}$, the frozen Teacher $\mu_{\mathrm{real}}$, and a trainable critic $\mu_{\mathrm{fake}}$ sharing the mHC-UNet backbone. The Student generates $x'_0 = G_{\phi}([\sigma_g z', C_{\mathrm{hifu}}], \sigma_g)$ where $z' \sim \mathcal{N}(0,I)$ and $\sigma_g$ is the generation noise scale and we align the Student's denoising field with the Teacher's prior. For a noisy state $x'_t = x'_0 + \sigma_t \epsilon$ perturbed at a sampled noise level $\sigma_t$ within the EDM distribution, both denoisers are evaluated:
\begin{equation}
\hat{x}^{\,\mathrm{real}}_0 = \mu_{\mathrm{real}}([x'_t, C_{\mathrm{hifu}}], \sigma_t),
\qquad
\hat{x}^{\,\mathrm{fake}}_0 = \mu_{\mathrm{fake}}([x'_t, C_{\mathrm{hifu}}], \sigma_t).
\end{equation}
The update direction $g$ is computed as:
\begin{equation}
w = \mathrm{mean}_{c,p}\!\left[\,\left|x'_0 - \hat{x}^{\,\mathrm{real}}_0\right|\,\right], \qquad
g = \frac{\hat{x}^{\,\mathrm{fake}}_0 - \hat{x}^{\,\mathrm{real}}_0}{w + \epsilon}.
\end{equation}
where $\epsilon=10^{-6}$ is a small constant for numerical stability. To ensure stability, the Student and critic are optimized via decoupled gradient flows:
\begin{equation}
\begin{aligned}
\mathcal{L}_{\mathrm{DMD}} &= \frac{1}{2}\left\|x'_0 - \mathrm{sg}\!\left(x'_0 - g\right)\right\|_2^2, \\
\mathcal{L}_{\mathrm{critic}} &= \mathbb{E}_{t}\!\left[ \left(\frac{1}{\sigma_t^2} + \frac{1}{\sigma_{\mathrm{data}}^2}\right) \left\|\mu_{\mathrm{fake}}([x'_t, C_{\mathrm{hifu}}], \sigma_t) - x'_0\right\|_2^2 \right],
\end{aligned}
\end{equation}
where $\mathrm{sg}(\cdot)$ denotes the stop-gradient operation. To prevent mode collapse in distillation procedure, we apply a supervised anchor $\mathcal{L}_{\mathrm{reg}} = \mathrm{LPIPS}(\hat{x}_0, x'_0)$ on a reference generation, and the final Student objective is $\mathcal{L}_{G} = \mathcal{L}_{\mathrm{DMD}} + \lambda_{\mathrm{reg}}\,\mathcal{L}_{\mathrm{reg}}$.

\section{Experiments}
\subsection{Datasets and Implementation}
\label{sec:setup}

\begin{figure}[b]
\centering
% 将 width 调整为 0.8\textwidth（或更小，如 0.7\textwidth）即可实现等比缩小
\includegraphics[width=0.7\textwidth]{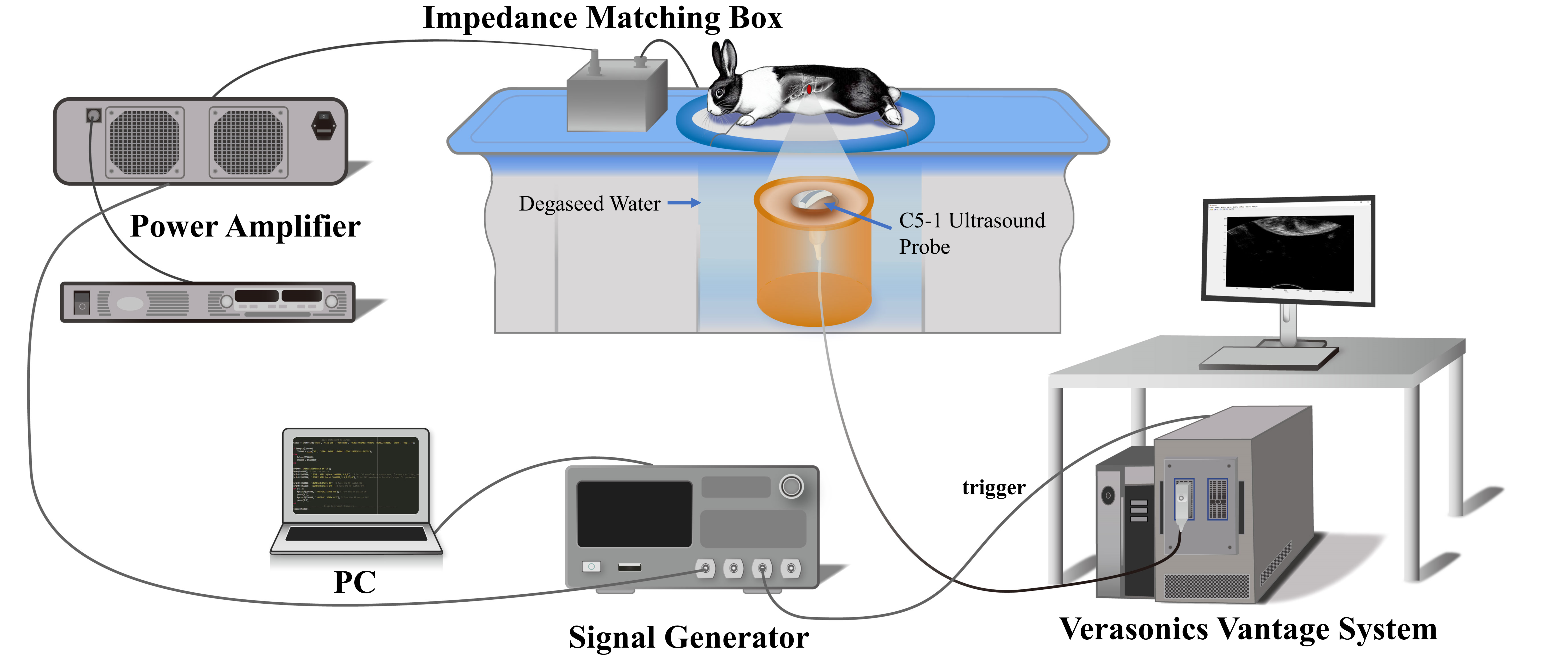}
\caption{Schematic illustration of the experimental setup. A clinical-grade HIFU therapy system is integrated with a 256-channel ultrasound research platform.}
\label{fig:exp_device}
\end{figure}

To evaluate the proposed framework, we constructed a large-scale dataset using a clinically equivalent HIFU therapy system (Focused Ultrasound Tumor Therapeutic System, JC200, Chongqing Haifu Medical Technology Co., Ltd.) integrated with a Verasonics Vantage 256 ultrasound research platform (Verasonics Inc., USA). The HIFU waveform was generated by a PC-controlled RIGOL DG4202 signal generator as a continuous 0.97 MHz signal, amplified, and delivered through an electrical matching box to the HIFU transducer. As illustrated in Fig.~\ref{fig:exp_device}, the HIFU transducer (0.97 MHz) was configured coaxially with an imaging probe to ensure aligned visualization and consistent monitoring of the focal region. The therapeutic transducer had a 0.97 MHz central frequency, 220 mm outer diameter, and 80 mm inner diameter, enabling coaxial placement of the imaging transducer.

\textbf{Data Acquisition and Protocol.} A consistent imaging depth of 135--140 mm was maintained across all imaging techniques. We implemented three clinically established beamforming strategies, i.e., Diverging Wave, Wide Beam, and Line Scan imaging, across four power levels (123 W--277 W). Paired corrupted and clean frames were acquired using a pulse-width modulation protocol (200 ms cycles), yielding 50 labeled paired samples per session. The final dataset comprises 17,464 image pairs, acquired at $512 \times 512$ and spatially resized to $256 \times 256$ for training, from ex vivo bovine and porcine tissues and in vivo rabbit models. To ensure independence, data were split at the subject/session level (7:1:2), yielding 12,225 training, 1,746 validation, and 3,493 test pairs. No session contributed frames to more than one split.

% \textbf{Data Acquisition and Protocol.} We implemented three clinically established beamforming strategies, namely, Diverging Wave, Wide Beam, and Line Scan imaging, across four power levels (123 W to 277 W). Paired corrupted and clean frames were acquired using a pulse width modulation protocol (200 ms cycles), yielding 50 labeled paired samples per session. The final dataset comprises 17,464 image pairs acquired from ex vivo bovine and porcine tissues, as well as in vivo rabbit models. These images were initially captured at a resolution of $512 \times 512$ and spatially resized to $256 \times 256$ for training. To ensure data independence, the dataset was split at the subject level (7:1:2), yielding 12,225 training, 1,746 validation, and 3,493 test pairs.

\textbf{Architecture and Training Details.} Both teacher and student models utilize an mHC-UNet backbone (128 channels, $n=4$, multipliers: $[2, 2, 2]$) coupled with an interference-aware encoder featuring five NAF blocks \cite{chen2022simple}. We adopt a two-stage training strategy: the encoder is jointly pre-trained alongside the teacher and then frozen during distillation to provide stable feature guidance. EDM parameters are set as $\sigma_{min}=0.002, \sigma_{max}=80, \sigma_{data}=0.5$. The teacher underwent 200 epochs of training (batch size 16, LR $1 \times 10^{-4}$), and the student used the same teacher configuration to ensure manifold consistency for DMD. The implementation was executed using PyTorch mixed precision on an NVIDIA H100, requiring $\sim$24 hours for end-to-end training. For fair comparison, all learning-based baselines followed the same preprocessing pipeline and subject/session-level splits, and supervised baselines were trained with identical restoration losses.

\subsection{Performance Evaluation and Comparison}
\label{sec:results}

We compare mHC-Diff against SOTA baselines, including signal processing methods (FRPCA) operating on raw RF signal, and deep learning architectures (FUS-Net, SwinUnet, and HIFU-Diff) in the image domain. We evaluate restoration quality using PSNR and SSIM (reported as mean with 95\% bootstrap confidence intervals).

\textbf{Quantitative Evaluation.} As shown in Table~\ref{tab:results}, mHC-Diff consistently outperforms all baselines across all beamforming modes. Specifically, in line-scan imaging, it yields a $27.29$ dB PSNR, surpassing the strongest baseline (SwinUnet) by $2.05$ dB. This superiority persists in more challenging wide-beam and diverging-wave scenarios, with narrow 95\% confidence intervals (via 1,000 bootstrap resamples) confirming its statistical stability. While FUS-Net and SwinUnet show moderate suppression, their lower SSIM indicates an inability to fully preserve fine anatomical details. SwinUnet benefits from global attention, but lacks an interference-conditioned prior for lesion fidelity. Notably, standard HIFU-Diff fails in the image domain as the absence of RF phase information in B-mode data creates nonlinear coupling between artifacts and anatomy, necessitating the robust priors of mHC-Diff for reliable disentanglement.

\begin{table}[b]
\centering
\caption{Quantitative comparison of different methods across three imaging modalities: Line Scan, Wide Beam, and Diverging Wave. Results are presented as Mean (95\% CI). Best results are highlighted in bold.}
\label{tab:results}
\resizebox{\textwidth}{!}{
\begin{tabular}{l c c c c c c}
\toprule
\multirow{2}{*}{Method} & \multicolumn{2}{c}{Line Scan} & \multicolumn{2}{c}{Wide Beam} & \multicolumn{2}{c}{Diverging Wave} \\
\cmidrule(lr){2-3} \cmidrule(lr){4-5} \cmidrule(lr){6-7}
 & PSNR $\uparrow$ & SSIM $\uparrow$ & PSNR $\uparrow$ & SSIM $\uparrow$ & PSNR $\uparrow$ & SSIM $\uparrow$ \\
\midrule
FUS-Net \cite{lee2021fus}    & 23.83 (23.66-24.00) & 0.812 (0.809-0.814) & 23.82 (23.68-23.96) & 0.809 (0.806-0.812) & 22.82 (22.63-22.99) & 0.805 (0.802-0.808) \\
HIFU-Diff \cite{cai2024suppressing} &  8.84  (8.50-9.16)  & 0.587 (0.576-0.598) &  7.31  (7.02-7.58)  & 0.535 (0.524-0.545) &  7.65  (7.35-7.96)  & 0.537 (0.528-0.548) \\
SwinUnet \cite{cao2022swin}  & 25.24 (25.08-25.41) & 0.838 (0.835-0.841) & 25.02 (24.88-25.17) & 0.838 (0.835-0.841) & 25.01 (24.83-25.20) & 0.836 (0.833-0.840) \\
FRPCA \cite{yang2024frequency}      & 15.59 (15.33-15.81) & 0.740 (0.732-0.747) & 10.77 (10.56-10.99) & 0.522 (0.515-0.529) & 13.44 (13.28-13.60) & 0.585 (0.581-0.589) \\
\midrule
\textbf{Ours (mHC-Diff)} & \textbf{27.29 (27.14-27.44)} & \textbf{0.858 (0.856-0.861)} & \textbf{26.20 (26.05-26.35)} & \textbf{0.858 (0.855-0.861)} & \textbf{25.49 (25.29-25.66)} & \textbf{0.850 (0.846-0.853)} \\
\bottomrule
\end{tabular}
}
\end{table}

\begin{figure}[t]
\centering
\includegraphics[width=0.8\textwidth]{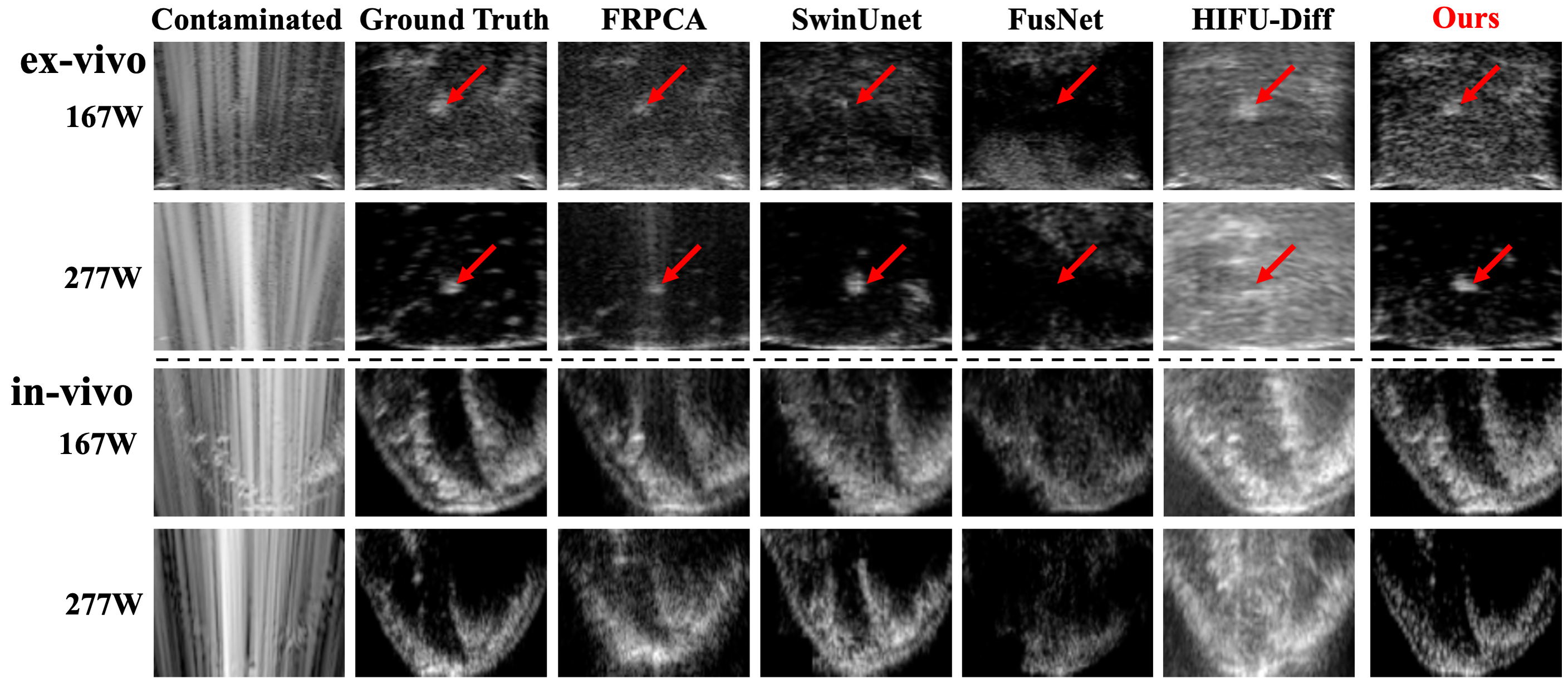}
\caption{\textbf{Qualitative comparison} across ex-vivo (top) and in-vivo (bottom) scenarios under varying power levels. \textbf{Red arrows} indicate critical HIFU-induced lesions.}
\label{fig:res} 
\end{figure}

\textbf{Qualitative Analysis.} 
As shown in Fig.~\ref{fig:res}, mHC-Diff achieves robust interference suppression while preserving anatomical structures across both ex vivo and in vivo settings. Conventional methods such as FRPCA and FUS-Net either fail to remove dense vertical streaks or suppress them at the cost of noticeable signal loss and contrast degradation. In contrast, mHC-Diff restores more natural speckle patterns and tissue contrast, even at the maximum power level (277~W). Crucially, HIFU-induced lesions (red arrows) are reconstructed with sharp edges and clear hyperechoic contrast, whereas baseline results are often blurred and partially merge lesion regions with background textures. With real-time inference at $\sim$20~FPS on a single NVIDIA RTX~4090, our framework provides reliable visual feedback for continuous ultrasound-guided HIFU procedures.

\begin{figure}[t]
\centering
% 【修改点】删除了 height=6cm，仅保留 width。
% 设置为 0.8\textwidth（或 0.85）可实现等比缩小并节省空间。
\includegraphics[width=0.8\textwidth]{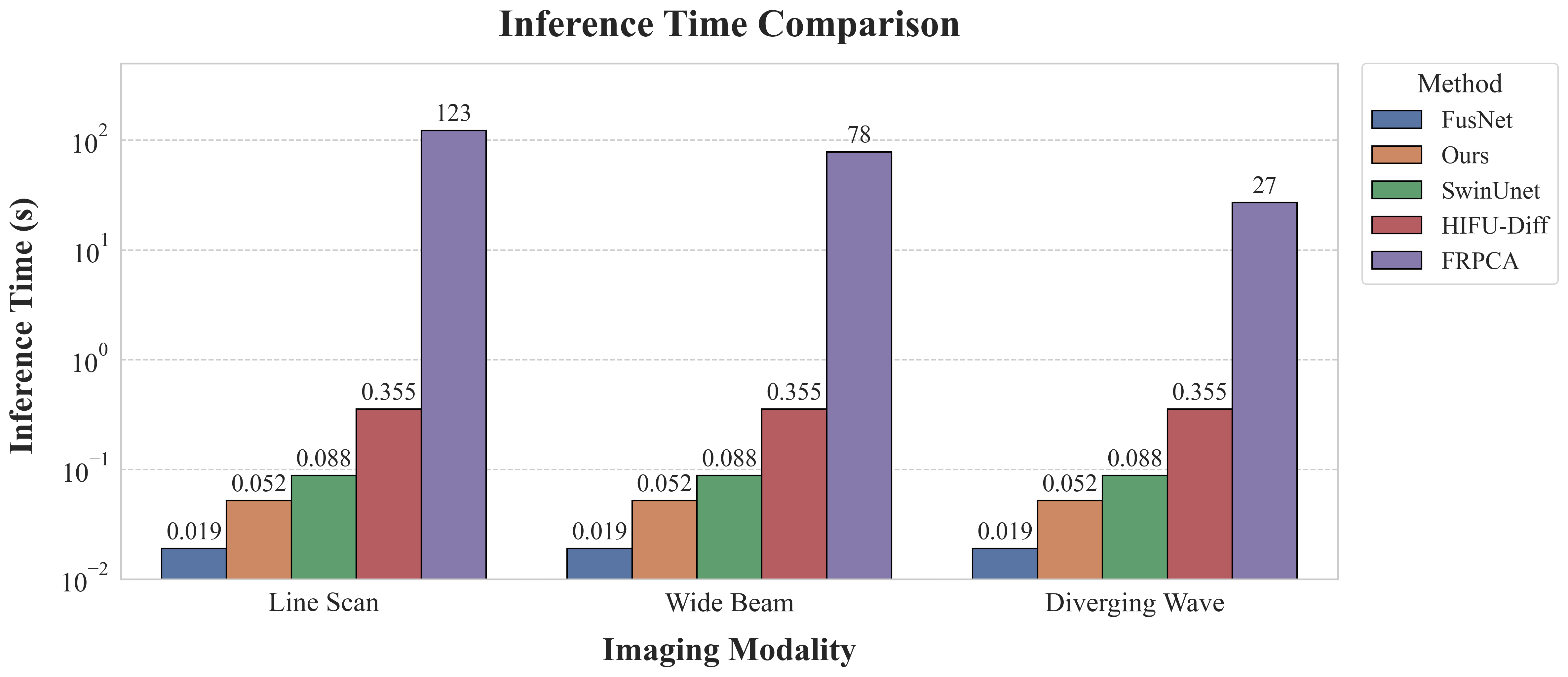}
\caption{Inference latency comparison. Our mHC-Diff maintains real-time throughput ($\approx$0.05 s) across modalities, significantly faster than HIFU-Diff and FRPCA baselines.}
\label{fig:time_comparison}
% \vspace{-0.4cm} 
\end{figure}

\subsection{Inference Efficiency}
\label{sec:efficiency}

Fig.~\ref{fig:time_comparison} compares computational efficiency across all modalities to assess intra-operative suitability. Traditional FRPCA exhibits prohibitive latency (e.g., 123 s/frame in line-scan mode), rendering it clinically incompatible. The standard HIFU-Diff requires 0.355 s/frame due to iterative sampling, creating a severe temporal bottleneck. Conversely, our one-step Student achieves a consistent 0.052 s/frame ($\sim$20~FPS) across all modalities. Although FUS-Net is the fastest (0.019 s/frame), its failure to preserve therapeutic features (Sec.~\ref{sec:results}) renders it clinically unreliable. SwinUnet (0.088 s/frame) remains significantly slower than our approach. By bypassing iterative sampling, our framework strikes an optimal balance between manifold integrity and real-time throughput. This enables seamless clinical integration for high-fidelity, low-latency visual feedback during active treatment.

\subsection{Ablation Study}
\label{sec:ablation}

Table~\ref{tab:ablation} quantifies the individual and synergistic contributions of the manifold-constrained Hyper-Connections (mHC) and Single-step Distillation (SD).

\begin{table}[!t]
\centering
\caption{Performance comparison and ablation study. The \textbf{Baseline} denotes the iterative EDM model without mHC or distillation, while \textbf{mHC-Diff} integrates both components. Best results are in \textbf{bold}.}
\label{tab:ablation}
\renewcommand{\arraystretch}{0.95}
\setlength{\tabcolsep}{2.6pt}
\resizebox{0.68\columnwidth}{!}{
\begin{tabular}{l c c c c c c}
\toprule
Method & PSNR $\uparrow$ & SSIM $\uparrow$ & MSE $\downarrow$ & MS-SSIM $\uparrow$ & LPIPS $\downarrow$ & Time (ms) $\downarrow$ \\
\midrule
FUS-Net \cite{lee2021fus} & 23.50 & 0.809 & 0.0054 & 0.6988 & 0.2068 & \textbf{19.0} \\
HIFU-Diff \cite{cai2024suppressing} & 7.93 & 0.553 & 0.2540 & 0.3779 & 0.2761 & 355.0 \\ 
SwinUnet \cite{cao2022swin} & 24.59 & 0.828 & 0.0040 & 0.8561 & 0.1539 & 88.0 \\
FRPCA \cite{yang2024frequency} & 14.33 & 0.648 & 0.0474 & 0.5930 & 0.1671 & $7.6 \times 10^4$ \\
\midrule
Baseline & 26.52 & 0.843 & 0.0022 & 0.8697 & 0.1431 & 499.0 \\ 
Baseline + mHC & \textbf{27.35} & 0.851 & \textbf{0.0018} & \textbf{0.8892} & 0.1426 & 592.5 \\
Baseline + Distill & 25.39 & 0.839 & 0.0029 & 0.8658 & 0.1499 & 38.5 \\
\midrule
\textbf{mHC-Diff (Ours)} & 26.65 & \textbf{0.856} & 0.0022 & 0.8875 & \textbf{0.1402} & 52.0 \\
\bottomrule
\end{tabular}
}
\end{table}

\textbf{Impact of Single-Step Distillation.} The standard iterative baseline (EDM teacher without mHC) achieves strong restoration (26.52~dB) but incurs high latency (499.0~ms). Distilling this teacher into a one-step student (Baseline+Distill) substantially reduces latency to 38.5~ms, at the cost of a PSNR drop to 25.39~dB. Despite this trade-off, the distilled model still outperforms supervised baselines such as SwinUnet (24.59~dB) and FUS-Net (23.50~dB). The teacher therefore serves as a training-time prior, not a deployed module. In contrast, directly applying HIFU-Diff to B-mode data yields very low PSNR (7.93~dB), suggesting that a dedicated image-domain prior is crucial for stable interference suppression.

\textbf{Synergy of mHC Routing and Distillation.} Integrating mHC to the iterative baseline improves the upper-bound restoration quality, where the mHC-equipped teacher achieves 27.35~dB PSNR and 0.0018 MSE. After distillation, our full model mHC-Diff recovers much of the fidelity typically lost in the one-step setting. Compared to Baseline+Distill (25.39~dB), mHC-Diff reaches 26.65~dB with the best structural and perceptual quality (0.856 SSIM; 0.1402 LPIPS). These results indicate that mHC routing effectively compensates for information loss introduced by distillation, yielding a favorable trade-off between image quality and real-time throughput for ultrasound-guided HIFU procedures.

% \begin{table}[t]
% \centering
% \caption{Ablation study evaluating the contribution of distillation and the mhc backbone. Best results are highlighted in bold.}
% \label{tab:ablation}
% \footnotesize % 1. 使用较小字号
% \setlength{\tabcolsep}{4pt} % 2. 缩减列间距，使表格更紧凑
% \resizebox{0.60\textwidth}{!}{ % 3. 将宽度从 \textwidth 降至 0.85 倍，可根据需要调整数值
% \begin{tabular}{l c c c c c}
% \toprule
% Method & PSNR $\uparrow$ & SSIM $\uparrow$ & MSE $\downarrow$ & MS-SSIM $\uparrow$ & LPIPS $\downarrow$ \\
% \midrule
% FUS-Net        & 23.50 & 0.809 & 0.0054 & 0.6988 & 0.2068 \\
% HIFU-Diff     &  7.93 & 0.553 & 0.2540 & 0.3779 & 0.2761 \\
% SwinUnet      & 24.59 & 0.828 & 0.0040 & 0.8561 & 0.1539 \\
% FRPCA         & 14.33 & 0.648 & 0.0474 & 0.5930 & 0.1671 \\
% \midrule
% HIFU-Teacher  & 26.52 & 0.843 & 0.0033 & \textbf{0.8897} & 0.1431 \\
% HIFU-Student  & 25.39 & 0.839 & \textbf{0.0029} & 0.8858 & 0.1499 \\
% HIFU-mHCTeacher  & 25.39 & 0.839 & \textbf{0.0029} & 0.8858 & 0.1799 \\
% \textbf{HIFU-mHCStudent (Ours)} & \textbf{26.65} & \textbf{0.856} & 0.0035 & 0.8875 & \textbf{0.1402} \\
% \bottomrule
% \end{tabular}
% }
% \end{table}
\FloatBarrier
\section{Conclusion}
\label{sec:conclusion}

We presented mHC-Diff, a two-stage framework for real-time HIFU interference suppression. By coupling manifold-constrained Hyper-Connections (mHC) with a one-step distillation paradigm, our approach effectively disentangles acoustic artifacts from anatomy, achieving 20 FPS inference. Extensive evaluations on a physically-consistent and clinically-representative corpus demonstrate superior reconstruction accuracy and structural fidelity over state-of-the-art benchmarks. The high-fidelity visualization of lesion boundaries in clinical scenarios underscores its potential for precise intraoperative guidance. Future work will focus on clinical hardware integration and broader cross-domain validation to further establish its robustness and therapeutic efficacy.
\section*{Acknowledgements}
This work was supported by the National Natural Science Foundation of China under grants 12304508 and 62402458, HKUST Research Project (Project No. 24251460C049), Hong Kong Innovation and Technology Commission (Project No. MHP/002/22), and Shenzhen Science and Technology Innovation Committee Fund (Project No. KCXFZ20230731094059008).
\section*{Disclosure of Interests}
The authors have no competing interests to declare.
\bibliography{references}

\begin{thebibliography}{10}
\providecommand{\url}[1]{\texttt{#1}}
\providecommand{\urlprefix}{URL }
\providecommand{\doi}[1]{https://doi.org/#1}

\bibitem{bond2017safety}
Bond, A.E., Shah, B.B., Huss, D.S., Dallapiazza, R.F., Warren, A., Ma, J.,
  Kassell, N.F., Elias, W.J.: Safety and efficacy of focused ultrasound
  thalamotomy for patients with medication-refractory, tremor-dominant
  parkinson disease: a randomized clinical trial. JAMA Neurology
  \textbf{74}(12),  1412--1418 (2017)

\bibitem{cai2024novel}
Cai, D., Yang, K., Liu, X., Xu, J., Ran, Y., Xu, Y., Zhou, X.: A novel
  diffusion-based deep learning model to suppress acoustic interference for
  real-time monitoring in ultrasound-guided hifu surgery. In: IEEE Ultrasonics,
  Ferroelectrics, and Frequency Control Joint Symposium (UFFC-JS). pp.~1--4.
  IEEE (2024). \doi{10.1109/UFFC-JS60046.2024.10793861}

\bibitem{cai2024suppressing}
Cai, D., Yang, K., Liu, X.B., Xu, J., Ran, Y., Xu, Y., Zhou, X.: Suppressing
  the hifu interference in ultrasound guiding images with a diffusion-based
  deep learning model. Computer Methods and Programs in Biomedicine
  \textbf{254},  108304 (2024)

\bibitem{cao2022swin}
Cao, H., Wang, Y., Chen, J., Jiang, D., Zhang, X., Tian, Q., Wang, M.:
  Swin-unet: Unet-like pure transformer for medical image segmentation. In:
  European Conference on Computer Vision (ECCV) Workshops. pp. 205--218.
  Springer Nature Switzerland, Cham (2022)

\bibitem{chen2022simple}
Chen, L., Chu, X., Zhang, X., Sun, J.: Simple baselines for image restoration.
  In: Proceedings of the European Conference on Computer Vision (ECCV). pp.
  17--33. Springer (2022)

\bibitem{chen2022ultrasound}
Chen, Z., Cheng, L., Zhang, W., He, W.: Ultrasound-guided thermal ablation for
  hyperparathyroidism: current status and prospects. International Journal of
  Hyperthermia  \textbf{39}(1),  466--474 (2022)

\bibitem{dhariwal2021diffusion}
Dhariwal, P., Nichol, A.: Diffusion models beat gans on image synthesis.
  Advances in neural information processing systems  \textbf{34},  8780--8794
  (2021)

\bibitem{dou2024long}
Dou, Y., Zhang, L., Liu, Y., He, M., Wang, Y., Wang, Z.: Long-term outcome and
  risk factors of reintervention after high intensity focused ultrasound
  ablation for uterine fibroids: a systematic review and meta-analysis.
  International Journal of Hyperthermia  \textbf{41}(1),  2299479 (2024)

\bibitem{esser2021taming}
Esser, P., Rombach, R., Ommer, B.: Taming transformers for high-resolution
  image synthesis. In: Proceedings of the IEEE/CVF conference on computer
  vision and pattern recognition. pp. 12873--12883 (2021)

\bibitem{ho2020denoising}
Ho, J., Jain, A., Abbeel, P.: Denoising diffusion probabilistic models.
  Advances in neural information processing systems  \textbf{33},  6840--6851
  (2020)

\bibitem{hynynen1993mri}
Hynynen, K., Darkazanli, A., Unger, E., Schenck, J.F.: Mri-guided noninvasive
  ultrasound surgery. Medical Physics  \textbf{20}(1),  107--115 (1993)

\bibitem{imankulov2018hifu}
Imankulov, S., Tuganbekov, T., Razbadauskas, A., Seidagaliyeva, Z.: Hifu
  treatment for fibroadenoma-a clinical study at national scientific research
  centre, astana, kazakhstan. J Pak Med Assoc  \textbf{68}(9),  1378--80 (2018)

\bibitem{karras2022elucidating}
Karras, T., Aittala, M., Aila, T., Laine, S.: Elucidating the design space of
  diffusion-based generative models. arXiv preprint arXiv:2206.00364  (2022).
  \doi{10.48550/arXiv.2206.00364}

\bibitem{lee2021fus}
Lee, S.A., Konofagou, E.E.: Fus-net: U-net-based fus interference filtering.
  IEEE transactions on medical imaging  \textbf{41}(4),  915--924 (2021)

\bibitem{liu2025machine}
Liu, Z., Liu, Z., Wang, Y., Wan, X., Huang, X.: Machine learning-based
  predictive analysis of energy efficiency factors necessary for the hifu
  treatment of adenomyosis. Frontiers in Physiology  \textbf{16},  1602866
  (2025). \doi{10.3389/fphys.2025.1602866}

\bibitem{niu2024acdmsr}
Niu, A., Pham, T.X., Zhang, K., Sun, J., Zhu, Y., Yan, Q., Kweon, I.S., Zhang,
  Y.: Acdmsr: Accelerated conditional diffusion models for single image
  super-resolution. IEEE Transactions on Broadcasting  \textbf{70}(2),
  492--504 (2024)

\bibitem{payen2023passive}
Payen, T., Crouzet, S., Guillen, N., Chen, Y., Chapelon, J.Y., Lafon, C.,
  Catheline, S.: Passive elastography for clinical hifu lesion detection. IEEE
  Transactions on Medical Imaging  \textbf{43}(4),  1594--1604 (2023)

\bibitem{saharia2022image}
Saharia, C., Ho, J., Chan, W., Salimans, T., Fleet, D.J., Norouzi, M.: Image
  super-resolution via iterative refinement. IEEE transactions on pattern
  analysis and machine intelligence  \textbf{45}(4),  4713--4726 (2022)

\bibitem{shen2022golay}
Shen, C.C., Lin, R.C., Wu, N.H.: Golay-encoded ultrasound monitoring of
  simultaneous high-intensity focused ultrasound treatment: a phantom study.
  IEEE Transactions on Ultrasonics, Ferroelectrics, and Frequency Control
  \textbf{69}(4),  1370--1381 (2022)

\bibitem{shen2024ultrasound}
Shen, C.C., Wu, N.H.: Ultrasound monitoring of simultaneous high-intensity
  focused ultrasound (hifu) therapy using minimum-peak-sidelobe coded
  excitation. Ultrasonics  \textbf{138},  107224 (2024)

\bibitem{song2014correspondence}
Song, J.H., Chang, J.H.: Correspondence-an effective pulse sequence for
  simultaneous hifu insonation and monitoring. IEEE Transactions on
  Ultrasonics, Ferroelectrics, and Frequency Control  \textbf{61}(9),
  1580--1587 (2014)

\bibitem{xie2025mhc}
Xie, Z., Wei, Y., Cao, H.b., Zhao, C., Deng, C., Li, J., Dai, D., Gao, H.,
  Chang, J., Yu, K., et~al.: mhc: Manifold-constrained hyper-connections. arXiv
  preprint arXiv:2512.24880  (2025)

\bibitem{yang2024suppressing}
Yang, K., Li, Q., Liu, H., Zeng, Q., Cai, D., Xu, J., Zhou, Y., Tsui, P.H.,
  Zhou, X.: Suppressing hifu interference in ultrasound images using 1d
  u-net-based neural networks. Physics in Medicine \& Biology  \textbf{69}(7),
  075006 (2024)

\bibitem{yang2024frequency}
Yang, K., Li, Q., Xu, J., Tang, M.X., Wang, Z., Tsui, P.H., Zhou, X.:
  Frequency-domain robust pca for real-time monitoring of hifu treatment. IEEE
  Transactions on Medical Imaging  \textbf{43}(8),  3001--3012 (2024)

\bibitem{yin2024one}
Yin, T., Gharbi, M., Zhang, R., Shechtman, E., Durand, F., Freeman, W.T., Park,
  T.: One-step diffusion with distribution matching distillation. arXiv
  preprint arXiv:2311.18828  (2024). \doi{10.48550/arXiv.2311.18828}

\bibitem{zhang2021designing}
Zhang, K., Liang, J., Van~Gool, L., Timofte, R.: Designing a practical
  degradation model for deep blind image super-resolution. In: Proceedings of
  the IEEE/CVF international conference on computer vision. pp. 4791--4800
  (2021)

\end{thebibliography}
\end{document}